\pdfoutput=1
\documentclass[11pt]{article}
\pdfoutput=1
\usepackage{ACL2023}

\usepackage{times}
\usepackage{latexsym}

\usepackage[T1]{fontenc}
\usepackage[utf8]{inputenc}

\usepackage{microtype}

\usepackage{inconsolata}

\usepackage{graphicx}
\usepackage{multirow}
\usepackage{booktabs}
\usepackage{doi}
\usepackage{url}
\usepackage{hyperref}
\usepackage{amssymb}
\usepackage{amsmath}
\usepackage{algorithm}
\usepackage{algorithmicx}  
\usepackage{algpseudocode}

\title{Leveraging Denoised Abstract Meaning Representation for Grammatical Error Correction}

\author{
Hejing Cao\textsuperscript{\rm 1,2}, 
Dongyan Zhao\textsuperscript{\rm 1,2}\thanks{\quad Corresponding author: Dongyan Zhao.}\\
$^1$ Wangxuan Institute of Computer Technology, Peking University\\
$^2$ Center for Data Science, Peking University \\
 \texttt {\{caohejing,zhaody\}@pku.edu.cn}\\ 
}

\begin{document}
\maketitle
\begin{abstract}
Grammatical Error Correction (GEC) is the task of correcting errorful sentences into grammatically correct, semantically consistent, and coherent sentences. Popular GEC models either use large-scale synthetic corpora or use a large number of human-designed rules. The former is costly to train, while the latter requires quite a lot of human expertise. In recent years, AMR, a semantic representation framework, has been widely used by many natural language tasks due to its completeness and flexibility. A non-negligible concern is that AMRs of grammatically incorrect sentences may not be exactly reliable. In this paper, we propose the AMR-GEC, a seq-to-seq model that incorporates denoised AMR as additional knowledge. Specifically, We design a semantic aggregated GEC model and explore denoising methods to get AMRs more reliable. Experiments on the BEA-2019 shared task and the CoNLL-2014 shared task have shown that AMR-GEC performs comparably to a set of strong baselines with a large number of synthetic data. Compared with the T5 model with synthetic data, AMR-GEC can reduce the training time by 32\% while inference time is comparable. To the best of our knowledge, we are the first to incorporate AMR for grammatical error correction.

\end{abstract}

\section{Introduction}

Nowadays, high performance of grammatical error correction model mainly depends on data augmentation \citep{kiyono-etal-2019-empirical, grundkiewicz-etal-2019-neural, raffel2020exploring, wan2021syntax, wu2022spelling, zhang-etal-2022-syngec}. According to the type of additional information, grammatical error correction models can be divided into data-enhanced models and knowledge-enhanced models. Data-enhanced models require millions of synthetic data, which is  obtained by back-translation or directly adding noise. Training on these synthetic datasets is very time-consuming, which is unacceptable in some application scenarios. Knowledge-enhanced model is to artificially design a large number of grammatical rule templates, and add the templates as external knowledge to GEC model. This external knowledge is language-dependent and it requires the intervention of human grammar experts. 
% When we design a grammatical error correction model for a new language, we still need to spend a lot of efforts on designing a set of grammatical rule templates for that language.

\begin{figure}
    \centering
    \includegraphics[width=0.3\textwidth]{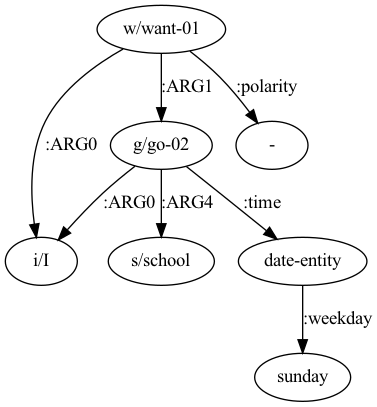}
    \caption{AMR of sentence "\emph{I don't want to go to school on Sunday.}"}
    \label{fig:example}
\end{figure}

Abstract Meaning Representation (AMR) is a type of rooted, labeled graph which contains semantic structures with fine-grained node and edge types. AMR breaks through the limitations of the traditional syntax tree structure and supports reentrancy. Figure \ref{fig:example} is a graph of sentence "\emph{I don't want to go to school on Sunday.}". In AMR, \emph{:arg0} is typically the agent, \emph{:arg1} is typically the patient, and other arguments do not have standard definitions and may vary with the verb being annotated. Negative meaning is denoted as "\emph{-}". Special keywords such as entity types, quantities and logical conjunctions are supported by AMR. AMR obtains a simple representation from natural language sentence and it is suitable for GEC as extra knowledge.

A non-negligible concern is that AMRs of errorful sentences may not be exactly reliable. If these AMRs with errors are directly introduced into the GEC model as additional information, it may confuse the model. We use a pre-trained AMR parser to predict AMR of erroneous sentences and corrected sentences separately on the BEA-19 development set. If two AMRs are completely consistent, we assume that the AMR of errorful sentences is reliable. After statistical analysis, we found that about half of the graphs are reliable. 
% The proportion of reliable graphs is quite large.
% Since a large part of the error types of GEC is distributed in preposition, singular/plural, and tense, which are dropped by AMR,  AMRs corresponding to this part of the error sentence are reliable.

We designed a denoising semantic aggregated grammatical error correction model. Specifically, we added a graph aggregation encoder based on a sequence-to-sequence model. The graph encoder aims to update the representation of the sequence encoder by AMR semantic structure. Besides, we designed two mask strategies to reduce the dependence on the model graph information. We designed these mask strategies by granularity: node/edge level mask and subgraph level mask. Experiments have proved that the denoising semantic aggregated grammatical error correction model significantly improved the error correction accuracy.

% In summary, the contribution of this paper is divided into three parts:

% \begin{enumerate}
%     \item We designed a semantic aggregated grammar error correction model (AMR-GEC), which is the first to add AMR as external knowledge to the grammar error correction task;
%     % \item We designed a graph denoise training method, which can reduce the information dependence of the model on the semantic graph of the wrong sentence and has achieved very good results.
%     \item The F score on the BEA-2019 restricted track reached 68.2\%, but the training speed is two times higher than that of the general model.
% \end{enumerate}

\section{Related works}

\textbf{Data-enhanced GEC models.} Lots of works have found their way to incorporating additional data into GEC model. \citet{kaneko-etal-2020-encoder} uses a pre-trained mask language model in grammatical error correction by using the output of BERT as additional features in the GEC model. \citet{kiyono-etal-2019-empirical} and \citet{grundkiewicz-etal-2019-neural} explore methods of how to generate and use the synthetic data and make use of Gigaword to construct hundreds of millions of parallel sentence pairs. Some works (\citealp{katsumata-komachi-2020-stronger}, \citealp{pajak2021grammatical}, \citealp{rothe-etal-2021-simple}) give a strong baseline by finetuning BART (\citealp{lewis-etal-2020-bart}), T5 (\citealp{raffel2020exploring}) on a GEC corpus. \citet{malmi-etal-2019-encode} casts GEC as a text editing task. \citet{zhao-etal-2019-improving} and \citet{panthaplackel2021copy} propose a copy-augmented architecture
for the GEC task by copying the unchanged
words and spans.

\noindent \textbf{Knowledge-enhanced GEC models.} \citet{wan2021syntax} use dependency tree as syntactic knowledge to guide the GEC model. \citet{wu2022spelling} adds part-of-speech features and semantic class features to enhance the GEC model. \citet{omelianchuk-etal-2020-gector} design thousands of custom token-level transformations to map input tokens to target corrections.  \citet{lai-etal-2022-type} proposes a multi-stage error correction model based on the previous model. 

% AMR is the most popular semantic representation framework in the NLP community and lots of downstream tasks benefit from it.
\noindent \textbf{Applications of AMR.} 
\citet{song-etal-2019-semantic} and  \citet{li-flanigan-2022-improving} incorporate AMR in neural machine translation. \citet{bonial-etal-2020-dialogue} makes use of AMR by abstracting the propositional content of an utterance in dialogue. \citet{xu2021dynamic} constructs a dynamic semantic graph employing AMR to cope with Multi-hop QA problems.
% \citet{dohare2017text} conducts text summarization by extracting and summarizing AMR.

% \noindent \textbf{Graph Neural Networks} Graph neural networks (GNN) have been proven effective in dealing with unstructured data problems. Graph structure contains rich information, which is very effective when dealing with unstructured data. The core of GNN is to pass information through the relationship between nodes. In recent years, graph convolutional networks (GCNs) and gated graph networks (GANs) have achieved success in various fields. GCNs is based on spectral theory. We perform Laplace transform to adjacency matrix A to define graph convolution. By choosing different convolution kernels, we can derive many variants of graph convolutional networks. E.g, SCNN \citep{defferrard2016convolutional} directly use learnable parameters as kernel, ChebNet  \citep{estrach2014spectral} use Chebyshev polynomials to reduce the amount of parameters, GCN \citep{kipf2016semi} further simplifies ChebyNet by using only first-order Chebyshev polynomials etc. GANs apply attention mechanism to graph convolution and propose a graph attention network to attent the relationships between nodes. GAT \citep{velivckovic2017graph} first applies attention mechanism to graph model, and GaAN \citep{zhang2018gaan} changes the attention method and enhances the ability to obtain information.

\section{Model}
\begin{figure}[htbp]
    \centering
    \includegraphics[width=0.4\textwidth]{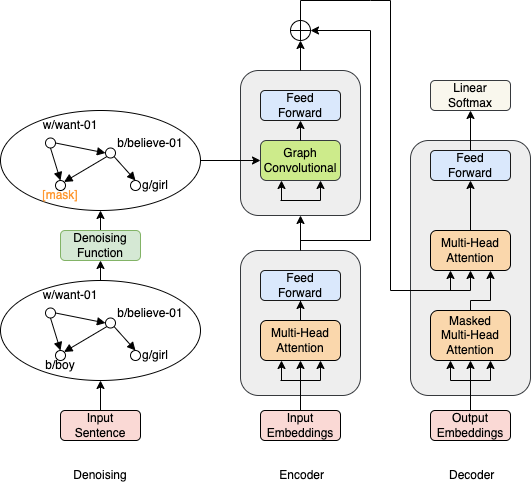}
    \caption{Denoising Semantic Aggregated GEC Model}
    \label{tab:amr-gec}
\end{figure}

% \subsection{Baseline}
We add a graph encoder based on Transformer to aggregate denoised semantic information. The architecture of AMR-GEC is shown on Figure \ref{tab:amr-gec}. 

\subsection{Semantic Aggregated Encoder}

Transformer is an attention-based encoder-decoder model, where the encoder encodes the input sentence into a context vector, and the decoder converts the context vector into an output sentence. Formally, we denote the tokens of the sentence is $T_n=\{t_1,t_2,...,t_n\}$. Vinilla encoder-decoder model works as follows:
\begin{align}
h_1,h_2,...,h_n & = {\rm Enc}(t_1,t_2,...,t_n) \\
y_1,y_2,...,y_m & = {\rm Dec}(h_1,h_2,...,h_n)
\end{align}

 We then designed a semantic graph encoder based on a graph attention network to incorporate semantic graph information. To preserve the information of the sequence encoder, we use a residual connection to combine the outputs of two encoders.
 \begin{align}
\hat{y}_1,\hat{y}_2,...,\hat{y}_m = {\rm GNN}(h_1,h_2,...,h_n) \\
y'_i = y_i \oplus \hat{y}_i,~~i=1,2,...,m
\end{align}
 % The hidden layer representation $h_1,h_2,...,h_n$ obtained by the sequence encoder can be regarded as a vector representation that incorporates sentence context information.

% Then, we map the spans to a meaning representation graph and make them aggregate neighborhood information by GAT.

% $$ \alpha_{i,j}^l=\frac{\exp(\sigma(f^l[Wh^l_i:W_ee_{i,j}:Wh^l_j]))}{\sum_{k\in N_i}\exp(\sigma(f^l[Wh^l_i:W_ee_{i,k}:Wh^l_k]))} $$
% $$ h^*_{i,k}=\sum_{j\in N_i}\alpha^l_{i,j}h^l_j~~~k=1,2,3,\cdots,M $$
% $$ h^{l+1}_i=\rm{Average(h^{*}_{i,1},~h^{*}_{i,2},~...,~h^{*}_{i,M})} $$
% where $h^l_i$ is the node i's hidden representation of layer l;  $e_{i,j}$ is the dependency between node i and node j.

\subsection{Denoising Function}

Masked Language Modeling (MLM) is a classic pre-trained model modeling method. The task of MLM is to mask some tokens with a special token \verb|mask| and train the model to recover them. This allows the model to handle both the left and right context of the masked token.  MLM can divided into five types: single word masking, phrase making, random span masking, entity masking, whole word masking.
% MLM is actually a cloze task, which restores the masked token according to the surrounding tokens of the masked token.

Referring to \citet{bai-etal-2022-graph}, we use the mask strategy on AMR. We used two ways to add masks: node/edge level mask and sub-graph level mask. Node/edge level mask refers to mapping the nodes/edges in the AMR graph using a noise function to generate a graph with noise. Sub-graph level mask means randomly removing subgraphs and replacing them with a mask label.

\subsection{Sequence-AMR Graph Construction}

In this section, we will show details about the graph encoder module. To preserve sequence information, we design a graph that fuses sequence and AMR. We first use the alignment tool JAMR to get the mapping from AMR node to sequence token. First connect the sequences through the special labels forward-label and backward-label respectively, and then map the edges of AMR to the sequence-AMR graph.

\begin{figure}[htbp]
    \centering
    \includegraphics[width=0.4\textwidth]{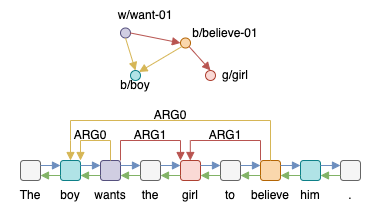}
    \caption{sequence-AMR graph}
    \label{fig:graph-c}
\end{figure}

\begin{algorithm}[htb]
  \caption{ Graph Construction}
  \label{alg:construction}
  \begin{algorithmic}[1]
    \Require AMR, sequence ($x_1$,$x_2$,...,$x_n$), Aligner
    \Ensure sequence-AMR graph
    \item 
    amr2seq = Aligner(sequence, AMR)
    \item graph= new Graph()
    \For{i=1 to n-1}
        \State AddEdge($x_i$, $x_{i+1}$, label-forward)
        \State AddEdge($x_{i+1}$, $x_{i}$, label-backward)
    \EndFor
    \For{edge in AMR.edges()}
        \State AddEdge(amr2seq[s], amr2seq[t], label)
    \EndFor
    \item return graph
  \end{algorithmic}
\end{algorithm}

\section{Experiments}

\subsection{Dataset}

% Currently, most research on grammatical error correction revolves around two public tasks, namely: 1) CoNLL2014; 2) BEA-2019. They both require developers to correct all types of grammatical errors.

\textbf{CoNLL-2014.} The CoNLL-2014 shared task test set contains 1,312 English sentences with error annotations by 2 expert annotators. Models are evaluated with M2 scorer (\citealp{dahlmeier-ng-2012-better}) which computes a span-based $F_{0.5}$-score.

\noindent \textbf{BEA-2019.} The BEA-2019 test set consists of 4477 sentences and the outputs are scored via ERRANT toolkit (\citealp{felice-etal-2016-automatic}, \citealp{bryant-etal-2017-automatic}). The released data are collected from Write \& Improve and LOCNESS dataset.

\subsection{Baseline Model} 

Following \citet{rothe-etal-2021-simple}, we use T5 as the baseline model for GEC.
% T5 is a Transformer-based text-to-text pre-trained model.
% which combines tasks such as translation, sentiment analysis, and text semantic similarity analysis in a unified framework.

\subsection{AMR Parsing and Alignment}

We adopt SPRING (\citealp{bevilacqua-etal-2021-one}) as our AMR parsing model. SPRING performs nearly state-of-the-art AMR parsing by linearizing AMR to sequence and converting text-to-amr task to seq-to-seq task. It obtained 84.5 Smatch F1 points on AMR 2.0 dataset.We use JAMR (\citealp{flanigan2014discriminative}) to align the AMRs to sentences. JAMR is an alignment-based AMR parsing model that finds a maximum spanning, connected subgraph as an optimization problem. We use the alignment for graph information aggregation.

\subsection{Others}

Our models were trained on a single GPU (GeForce GTX 1080), and our implementation
was based on publicly available code\footnote{\url{https://github.com/huggingface/transformers}}. we set the batch\_size to 6 and the learning\_rate to 2e-5. We use pytorch\_geometric\footnote{\url{https://github.com/pyg-team/pytorch_geometric}} to implement the semantic aggregated encoder. 

% We release our code at \url{https://github.com/hejingcao/AMR-GEC}.

\section{Results and Analysis}

\subsection{Results}

\begin{table*}[htbp]
\centering
\begin{tabular}{lccccccc}
\hline
\multicolumn{1}{c}{\multirow{2}{*}{\textbf{Models}}} & \multicolumn{1}{c}{\multirow{2}{*}{\textbf{Synthetic data}}} & \multicolumn{3}{c}{\textbf{BEA-test}} & \multicolumn{3}{c}{\textbf{CoNLL-14}} \\ \cline{3-8} 
\multicolumn{1}{c}{}                        & \multicolumn{1}{c}{}                             & P        & R       & $F_{0.5}$    & P        & R       & $F_{0.5}$    \\ \hline
\citet{katsumata-komachi-2020-stronger}     & -                                                & 68.3     & 57.1    & 65.6    & 69.3     & 45.0    & 62.6    \\
\citet{kiyono-etal-2019-empirical}          & \checkmark                                                & 69.5     & 59.4    & 64.2    & 67.9     & 44.1    & 61.3    \\
\citet{kaneko-etal-2020-encoder}            & \checkmark                                                & 67.1     & \textbf{61.0}    & 65.6    & 69.2     & 45.6    & 62.6    \\
\citet{rothe-etal-2021-simple}              & \checkmark                                                & -        & -       & 67.1    & -        & -       & 65.1    \\
\citet{omelianchuk-etal-2020-gector}        & \checkmark                                                & \textbf{79.2}     & 53.9    & \textbf{72.4}    & \textbf{77.5}     & 40.1    & \textbf{65.3}    \\
AMR-GEC                                     & -                                                & 71.5     & 58.3    & 68.4    & 70.2     & \textbf{48.3}    & 64.4    \\ \hline
\citet{katsumata-komachi-2020-stronger}     & -                                                & 68.8     & 57.1    & 66.1    & 69.9     & 45.1    & 63.0    \\
\citet{kiyono-etal-2019-empirical}          & \checkmark                                                & 74.7     & 56.7    & 70.2    & 67.3     & 44.0    & \textbf{67.9}    \\
\citet{omelianchuk-etal-2020-gector}        & \checkmark                                                & \textbf{79.4}     & \textbf{57.2}    & \textbf{73.7}    & \textbf{78.2}     & 41.5    & 66.5    \\
AMR-GEC                                     & -                                                & 73.5     & 55.9    & 69.1    & 70.3     & \textbf{48.2}    & 64.4    \\ \hline
\end{tabular}                        
\caption{Results of AMR-GEC. The first group shows the results of single models. The second group shows the results of ensemble models. The ERRANT for BEA-test and the $M^2$ score for CoNLL-14 (test) are reported. we simply rerank outputs by generation probabilities of single models.}
\label{tab:results}
\end{table*}

Table \ref{tab:results} shows the results of the BEA-test and CoNLL-2014 dataset. 1) Compared with the model without synthetic data, the single model of AMR-GEC is 2.8 points and 1.8 points higher in BEA-19 and CoNLL-14, respectively. Ensemble models give similar results. 2) Compared with models using synthetic data, AMR-GEC gives comparable or even higher F-score, except for GECToR \citep{omelianchuk-etal-2020-gector}, which uses both synthetic data and human knowledge. For example, our single model achieves 68.4 on BEA-19, higher than the models by \citet{kiyono-etal-2019-empirical}, \citet{kaneko-etal-2020-encoder}, and \citet{rothe-etal-2021-simple}. This shows that semantic graphs, as additional knowledge for GEC, have a comparative advantage over synthetic data.
Our ensemble model does not show significant improvements over the single model, probably because more optimal ensemble strategies are needed: averaging generation probabilities (\citealp{omelianchuk-etal-2020-gector}), ensemble editings (\citealp{pajak2021grammatical}), etc.

\subsection{Advantages of AMR}

\begin{table}[htbp]
\centering
\small
\begin{tabular}{lllllll}
\hline
\multicolumn{1}{c}{\multirow{2}{*}{Error   Type}} & \multicolumn{3}{c}{T5-GEC} & \multicolumn{3}{c}{AMR-GEC} \\ \cline{2-7} 
\multicolumn{1}{c}{}                              & P     & R     & $F_{0.5}$  & P      & R     & $F_{0.5}$  \\ \hline
PUNCT                                             & 79.8  & 49.4  & 71.0       & 78.7   & 72.9  & \textbf{77.4}       \\
DET                                               & 78.6  & 64.8  & 75.4       & 78.6   & 65.8  & \textbf{75.7}       \\
PREP                                              & 72.9  & 48.0  & 66.0       & 73.1   & 61.5  & \textbf{70.4}       \\
ORTH                                              & 84.6  & 55.7  & \textbf{76.7}       & 69.5   & 62.9  & 68.1       \\
SPELL                                             & 83.0  & 58.3  & \textbf{76.5}       & 80.9   & 61.9  & 76.2       \\ \hline
\end{tabular}
\caption{BEA-test scores for the top five error types, except for OTHER}
\label{tab:error}
\end{table}

We compared the most common error types in BEA-test (except for OTHER) between T5-GEC and AMR-GEC. As shown in Table \ref{tab:error}, the F-scores of PUNCT and PREP in AMR-GEC is 4-6 points higher than T5-GEC. AMR dropped prepositions, tense, and punctuation to obtain simple and base meanings, and exactly these error types are the most common errors in GEC scenarios. With such error ignored in AMR, sentences generated from AMR are more likely to get correct results.

Besides, graphs are good at solving the long sentence dependency problem. The pain point of the sequence model is that it is difficult to pay attention to long-distance dependent information. In AMR, associative concept nodes are explicitly connected with edges, making it easier for the model to focus on long-distance information.

% \subsection{Error type analysis}

\section{Ablation Study}

\subsection{Graph Neural Networks Ablation Results}

Graph neural networks have been proven effective in dealing with unstructured data problems. However, few studies have analyzed the effect of different GNN-encoded AMRs for natural language generation tasks. To study the differences of graph neural networks of  encoding AMR, we carry on a set of experiments. We select different graph encoders of GCN, GAT, and DeepGCN as variables, and conduct experiments on BEA-2019 dataset while ensuring the same amount of model parameters. We do not use the denoising method in this ablation study.

% We finetuned T5-base on BEA-2019 training set as our baseline, denoted as T5-GEC; the models using GCN, GAT, and DeepGCN as graph encoder were respectively denoted as AMR-GCN, AMR-GAT, and AMR-DeepGCN. Among them, the number of attention headers of GAT is 4, and the number of network layers of deepGCN is 4.

\begin{table}[htbp]
\centering
\small
\begin{tabular}{@{}lccc@{}}
\toprule
\textbf{Model}       & \textbf{P}     & \textbf{R}     & \textbf{$F_{0.5}$} \\ \midrule
T5-GEC      & 71.47 & 53.46 & 66.96 \\
AMR-GCN     & \textbf{72.95} & 52.17 & \textbf{67.57} \\
AMR-GAT     & 68.26 & \textbf{63.41} & 67.23 \\
AMR-DeepGCN & 66.34 & 62.57 & 65.55 \\ \bottomrule
\end{tabular}
\caption{Results on BEA-test with GCN, GAT, DeepGCN as AMR encoders}
\label{tab:gnn}
\end{table}

Table \ref{tab:gnn} shows the results of BEA-test with different graph encoders. We can draw these conclusions: 1) Even if the AMRs of the errorful sentences are not reliable, they still benefit GEC. Compared with T5-GEC, AMR-GCN and AMR-GAT are about 0.2 and 0.4 points higher respectively. This shows that the model makes use of the semantic information and connection relationship of reliable AMR. 2) AMR-GCN gives the best performance among the three models. When picking a graph encoder, the GCN model is sufficient to encode the semantic structure information of AMR. It is worth noting that GAT and DeepGCN have high recall value and low precision. In the grammatical error correction task, precision measures the error correction result. Generally speaking, precision is more important than recall. In the grammatical error correction task, most of the errors are local errors, and the semantic information required for grammatical error correction in AMR can be captured without a deeper graph convolution model.

\subsection{Denoise method ablation study}

\begin{table}[htbp]
\centering
\small
\begin{tabular}{@{}lccc@{}}
\toprule
\textbf{Model}       & \textbf{P}     & \textbf{R}     & \textbf{$F_{0.5}$} \\ \midrule
T5-GEC                   & 71.47 & 53.46 & 66.96 \\
AMR-GCN                  & 72.95 & 52.17 & 67.57 \\
AMR-GCN (node/edge) & \textbf{73.52} & 55.91 & \textbf{69.14} \\
AMR-GCN (subgraph)  & 72.12 & \textbf{57.60} & 68.60 \\ \bottomrule
\end{tabular}
\caption{Results on BEA-test with node/edge and subgraph denoising methods}
\label{tab:mask}
\end{table}

Table \ref{tab:mask} shows the results of BEA-test with node/edge and subgraph denoising methods. The node/edge level denoising strategy and the subgraph level denoising strategy increased by 1.57 and 1.03 points, respectively. Node level mask strategy performs better because the subgraph may mask too much information.

% \section{Case Study}

% \begin{figure*}
%     \centering
%     \includegraphics[width=0.6\textwidth]{figures/case.png}
%     \caption{AMR}
% \end{figure*}

\section{Conclusion}

In this paper, We propose a denoising  semantic aggregated grammatical error correction model, AMR-GEC, leveraging AMR as external knowledge to the GEC. We believe it gives a strong baseline for incorporating AMR in GEC.

\section*{Limitations}
% ACL 2023 requires all submissions to have a section titled ``Limitations'', for discussing the limitations of the paper as a complement to the discussion of strengths in the main text. This section should occur after the conclusion, but before the references. It will not count towards the page limit.
% The discussion of limitations is mandatory. Papers without a limitation section will be desk-rejected without review.

% While we are open to different types of limitations, just mentioning that a set of results have been shown for English only probably does not reflect what we expect. 
% Mentioning that the method works mostly for languages with limited morphology, like English, is a much better alternative.
% In addition, limitations such as low scalability to long text, the requirement of large GPU resources, or other things that inspire crucial further investigation are welcome.

In this paper, we leverage AMR to the GEC model as external knowledge, and achieve  a high F-score on single model. However, we do not use R2L reranking, model ensemble and other methods to ensemble single model and compare them with state-of-the-art ensemble models. Our aim is to provide a strong baseline for incorporating AMR in GEC, so it is easy to generalize AMR-GEC to ensemble models.

\section*{Ethics Statement}

The training corpora including the Lang-8, NUCLE and the BEA-2019 test data and CoNLL-2014 test data used for evaluating our framework are publicly available and don’t pose privacy issues. The algorithm that we propose does not introduce ethical or social bias.

% ToDo
% Scientific work published at ACL 2023 must comply with the ACL Ethics Policy.\footnote{\url{https://www.aclweb.org/portal/content/acl-code-ethics}} We encourage all authors to include an explicit ethics statement on the broader impact of the work, or other ethical considerations after the conclusion but before the references. The ethics statement will not count toward the page limit (8 pages for long, 4 pages for short papers).

\section*{Acknowledgements}
We would like to thank the anonymous reviewers
for their constructive comments. We would like to express appreciation to Yansong Feng for his  insightful suggestions on the algorithm framework. This
work was supported by the National Key Research
and Development Program of China (No.
2020AAA0106600).
% including
% John Chen, Henry S. Thompson and Donald Walker.
% Additional elements were taken from the formatting instructions of the \emph{International Joint Conference on Artificial Intelligence} and the \emph{Conference on Computer Vision and Pattern Recognition}.

\bibliography{anthology, custom}
% \bibliography{anthology}
\bibliographystyle{acl_natbib}

% \appendix

\end{document}